\documentclass[letterpaper]{article} 
\usepackage{aaai2026}  
\usepackage{times}  
\usepackage{helvet}  
\usepackage{courier}  
\usepackage[hyphens]{url}  
\usepackage{graphicx} 
\usepackage{natbib}  
\usepackage{caption} 
\usepackage{algorithm}
\usepackage{algorithmic}

\usepackage{amssymb}
\usepackage{amsmath}

\usepackage[pagewise]{lineno}

\usepackage{subcaption} 
\usepackage[export]{adjustbox} 
\usepackage{booktabs} 
\usepackage{multirow} 
\usepackage{makecell} 
\usepackage{arydshln} 
\usepackage{times}
\usepackage{helvet}
\usepackage{courier}
\usepackage{xcolor}

\usepackage{newfloat}
\usepackage{listings}
\DeclareCaptionStyle{ruled}{labelfont=normalfont,labelsep=colon,strut=off} 
\floatstyle{ruled}
\newfloat{listing}{tb}{lst}{}
\floatname{listing}{Listing}
\title{FedSEA-LLaMA: A Secure, Efficient and Adaptive Federated Splitting \\Framework for Large Language Models}
\author{
    Zishuai Zhang\textsuperscript{\rm 1,\rm 2},
    Hainan Zhang\textsuperscript{\rm 1,\rm 2}\thanks{Corresponding author},
    Weihua Li\textsuperscript{\rm 2},
    Qinnan Zhang\textsuperscript{\rm 1,\rm 2},
    Jin Dong\textsuperscript{\rm 1, \rm 3}\footnotemark[1],
    Yongxin Tong\textsuperscript{\rm 4}\footnotemark[1],
    Zhiming Zheng\textsuperscript{\rm 1,\rm 2}
}
\affiliations {
    \textsuperscript{\rm 1}Beijing Advanced Innovation Center for Future Blockchain and Privacy Computing, Beihang University, China\\
    \textsuperscript{\rm 2}School of Artificial Intelligence, Beihang University, China\\
    \textsuperscript{\rm 3}Beijing Academy of Blockchain and Edge Computing, China\\
    \textsuperscript{\rm 4}School of Computer Science and Engineering, Beihang University, China\\
    zhangzishuai@buaa.edu.cn, zhanghainan@buaa.edu.cn
}

\begin{document}

\maketitle

\begin{abstract}
Private data holds promise for improving LLMs due to its high quality, but its scattered distribution across data silos and the high computational demands of LLMs limit their deployment in federated environments. To address this, the transformer-based federated split models are proposed, which offload most model parameters to the server (or distributed clients) while retaining only a small portion on the client to ensure data privacy. Despite this design, they still face three challenges: 1) Peer-to-peer key encryption struggles to secure transmitted vectors effectively; 2) The auto-regressive nature of LLMs means that federated split learning can only train and infer sequentially, causing high communication overhead; 3) Fixed partition points lack adaptability to downstream tasks. In this paper, we introduce FedSEA-LLaMA, a Secure, Efficient, and Adaptive Federated splitting framework based on LLaMA2. First, we inject Gaussian noise into forward-pass hidden states to enable secure end-to-end vector transmission. Second, we employ attention-mask compression and KV cache collaboration to reduce communication costs, accelerating training and inference. Third, we allow users to dynamically adjust the partition points for input/output blocks based on specific task requirements. Experiments on natural language understanding, summarization, and conversational QA tasks show that FedSEA-LLaMA maintains performance comparable to centralized LLaMA2 and achieves up to 8× speedups in training and inference. Further analysis of privacy attacks and different partition points also demonstrates the effectiveness of FedSEA-LLaMA in security and adaptability. 
\end{abstract}

\begin{links}
    \link{Code}{https://github.com/TAP-LLM/SplitFedLLM/tree/main/FedSEA-LLaMA}
    \link{Extended version}{https://arxiv.org/abs/2505.15683}
\end{links}

\section{Introduction}
While the success of Large Language Models (LLMs) has largely stemmed from their ability to process vast amounts of public data~\cite{kaplan2020scaling}, the true potential of these models lies in their application to high-quality private data. Private datasets often contain more granular and specific insights about individuals or organizations, typically collected through more controlled and precise methodologies~\cite{yang2024federated}. Additionally, private data tend to be more timely, targeted, and unique, thus providing models with semantic contexts and environmental settings that more closely mirror real-world scenarios~\cite{ye2024openfedllm}. Therefore, effectively incorporating private data will further advance the performance of LLMs.

\begin{figure}[!t]
    \centering
    \includegraphics[width=\linewidth]{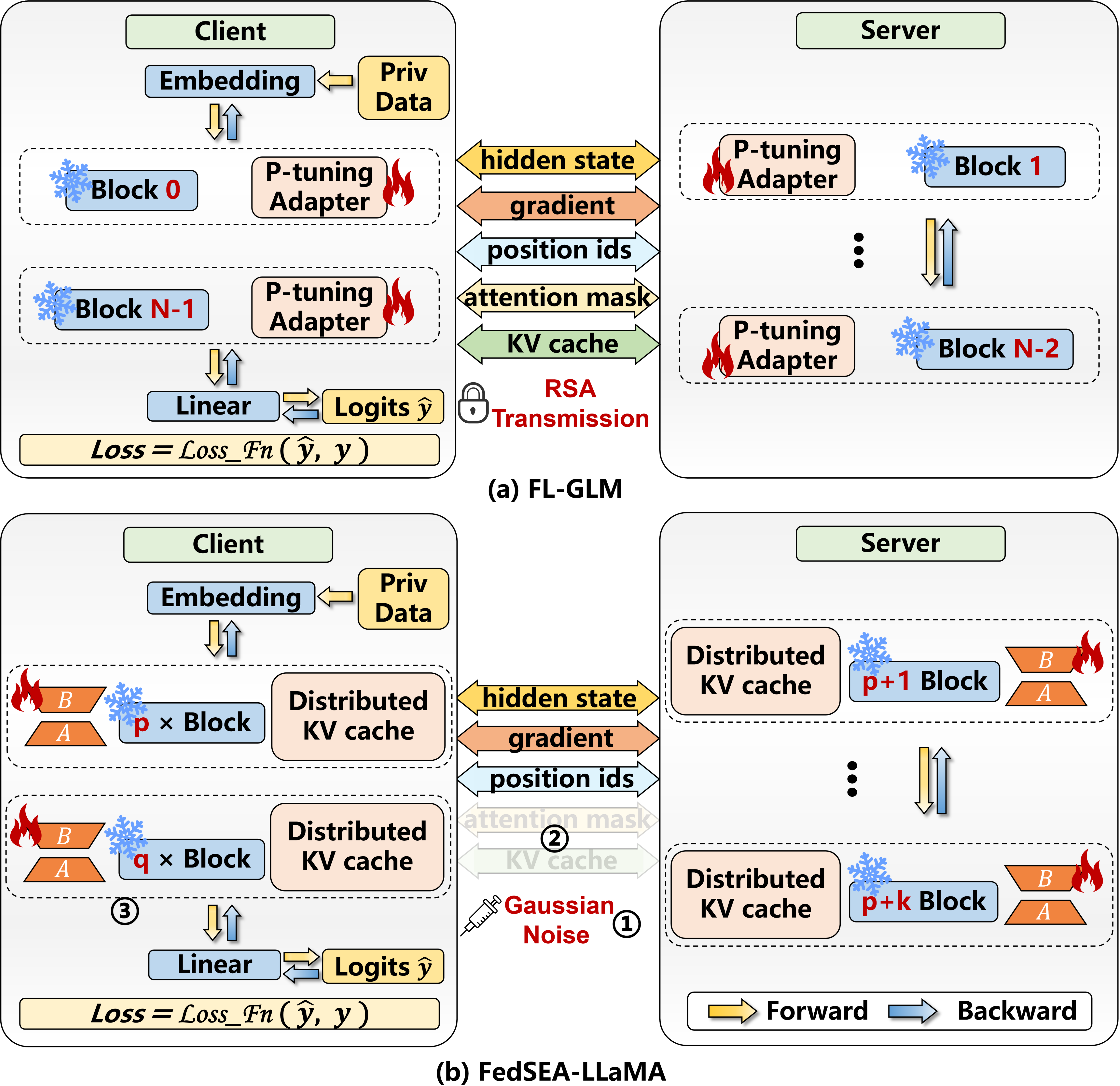}
    \caption{ Model architecture of FL-GLM and FedSEA-LLaMA.
    \textcircled{{\small 1}}FedSEA-LLaMA employs Gaussian noise addition to protect intermediate vectors. 
    \textcircled{{\small 2}}FedSEA-LLaMA minimizes the transmission of intermediate data and further accelerates long-context inference. 
    \textcircled{{\small 3}}FedSEA-LLaMA adaptively adjusts model partition points based on task requirements and system load.
   }
    \label{fig:compare}
\end{figure}

The distributed private data silos and the significant computational demands of LLMs pose substantial challenges for their deployment in federated environments. Private data is often isolated in local silos, such as mobile devices~\cite{ghimire2022recent}, enterprise servers~\cite{huang2024federated}, or medical institutions~\cite{yuan2024decentralized}, as centralized storage and processing of such data may raise privacy concerns and regulatory risks~\cite{chen2024federated}. Federated Learning (FL)~\cite{stremmel2021pretraining} offers a promising solution by enabling local training on user devices and only sending model parameters for aggregation. They often leverage Adapters~\cite{houlsby2019parameter} or LoRA~\cite{hu2022lora} as transmitted parameters, or employ collaborative training between large and small models~\cite{hinton2015distilling} to fine-tune LLMs. However, in extremely resource-constrained scenarios (e.g., less than 3.5GB of GPU memory), fine-tuning the LLMs remains infeasible. Therefore, how to effectively deploy LLMs in FL frameworks while ensuring privacy and scalability in resource-constrained scenarios remains an open research problem.

To address this, transformer-based federated splitting models~\cite{zheng2024safely,su2024titanic} have been proposed to adapt LLM fine-tuning to resource-constrained scenarios. Titanic~\cite{su2024titanic} handles limited server resources by partitioning LLM across clients and enabling collaborative training. FL-GLM~\cite{zheng2024safely} offloads most model parameters to a resource-ample server while retaining the input and output layer on the client. As shown in Figure~\ref{fig:compare}a, FL-GLM divides Transformers into three parts, i.e., the first input block and the last output block on the client, while the middle blocks on the server. The forward propagation (FP) path is client-to-server-to-client with peer-to-peer key encryption. After the client calculates loss, gradients backpropagate (BP) from client-to-server-to-client with the same encryption. Although FL-GLM provides a potentially viable solution for clients in resource-constrained scenarios, it still faces several limitations: \textbf{1) Privacy Leakage:} during forward and backward propagation, activation values and gradients from the input blocks are transmitted, which can be exploited to reconstruct original input data through reverse engineering~\cite{asnani2023reverse}. \textbf{2) Communication Cost:} LLMs are auto-regressive, requiring sequential token generation based on prior tokens. In federated split learning, this necessitates a full forward and backward pass with client-server communication for each token, greatly increasing communication rounds and bandwidth usage. \textbf{3) Lack of Adaptability: } Fixed partition points of FL-GLM lack flexibility to downstream tasks and hardware constraints, as different devices vary in their capacity to handle workloads and parameter sizes. 

In this paper, we introduce FedSEA-LLaMA, a Secure, Efficient, and Adaptive Federated split framework based on LLaMA2, as shown in Figure~\ref{fig:compare}b. To \textbf{Enable Secure Propagation}, we add Gaussian noise for the hidden states of input blocks and send these noised states to the server for further processing. Through both gradient analysis and experimental verification, this noise ensures privacy and security during subsequent training and inference stages. To \textbf{Reduce Communication Cost}, we introduce the attention-mask compression and collaborative KV cache mechanisms, minimizing overhead while preserving the integrity of auto-regressive inference. By synchronously maintaining cache states on both client and server, the originally cross-device transmission is transformed into dynamic attention-mask generation, reducing communication from megabytes (MB) to bytes (B). To \textbf{Achieve Adaptive Partition}, we design a dynamic partition strategy allowing users to flexibly adjust the number of input and output blocks according to specific downstream tasks and available computational resources. This adaptability enables an optimal balance between performance and efficiency.

Experiments on natural language understanding (NLU), summarization, and conversational QA show that FedSEA-LLaMA maintains performance comparable to centralized LLaMA2 and achieves up to 8× inference speedups. Through inference acceleration, the average token generation time of FedSEA-LLaMA has been reduced by 87.6\%, and the speed gain will further increase with the growing length of context. Additional analysis of privacy attacks and various partition points further highlights FedSEA-LLaMA's effectiveness in terms of security and adaptability.

The innovations in this paper are as follows:
\begin{itemize}
    \item We design a federated split framework tailored for resource-limited scenarios, featuring adaptive partitioning and client-side Gaussian noise injection to enable privacy-preserving training and inference of LLMs.
    
    \item We leverage collaborative KV caching and dynamic attention-mask generation to drastically reduce communication overhead from megabytes to bytes during auto-regressive inference, achieving up to 8x speedups.
    
    \item  Experimental results on three different tasks show that our FedSEA-LLaMA achieves performance comparable to centralized LLaMA2, and validate the security, efficiency, and adaptability of our framework.
\end{itemize}

\section{Related Works}
\begin{figure*}[t!]
    \centering
    \includegraphics[width=0.85\linewidth]{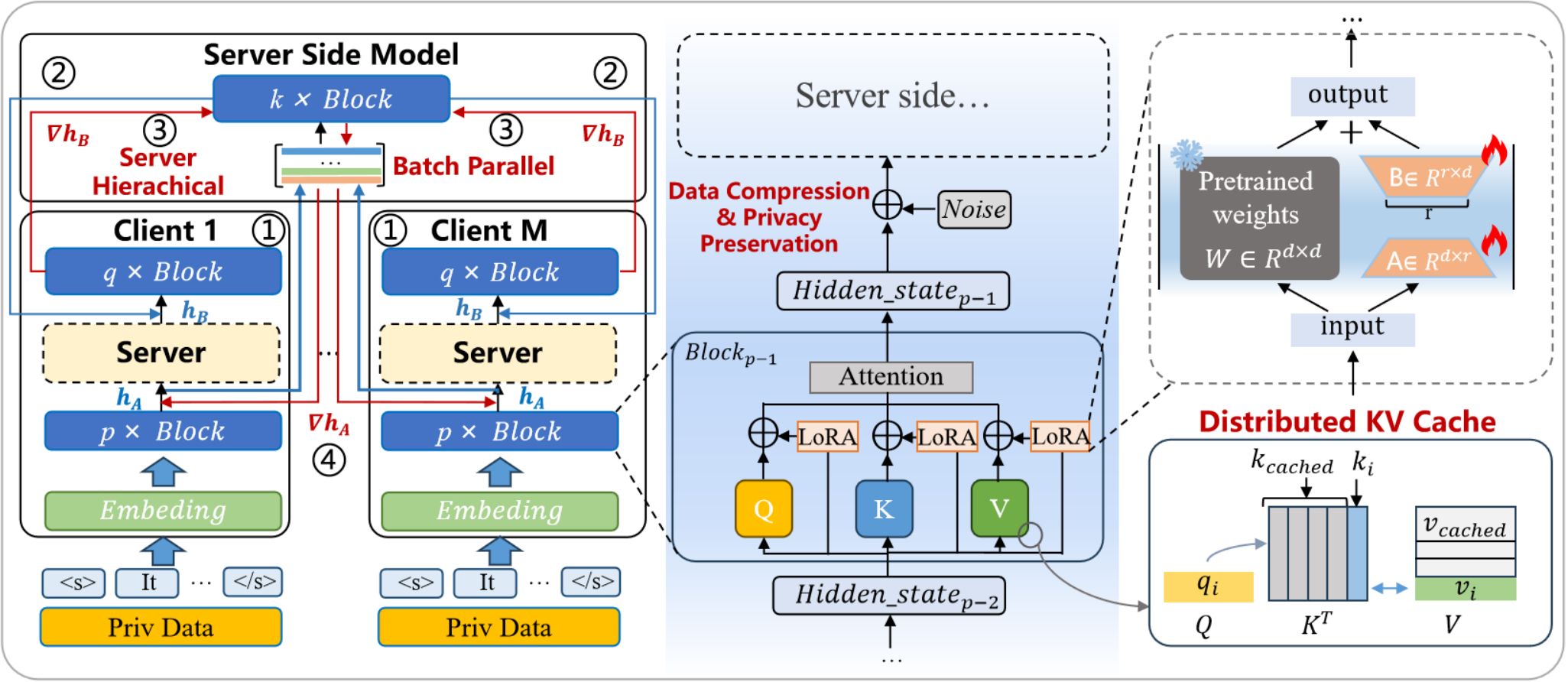}
    \caption{Overall Framework of FedSEA-LLaMA. Model Split with adaptive partition points and Gaussian noise on forward-passing vectors to preserve privacy for LoRA fine-tuning strategy, and achieving attention-mask compression and KV Cache collaborations to accelerate inference process. The local client stores the first ${p}$ blocks ($Blocks_A$) and the last ${q}$ blocks ($Blocks_C$), while the server stores the remaining ${k}$ blocks ($Blocks_B$).
    During training, \textcircled{{\small 1}} $h_A$ is transmitted to the server-side model $Blocks_B$; \textcircled{{\small 2}} $h_B$ is transmitted to the client-side model $Blocks_C$; \textcircled{{\small 3}} the gradient with respect to $h_B$ is backpropagated to the server; \textcircled{{\small 4}} the gradient with respect to $h_A$ is backpropagated to the client-side model.}
    \label{fig:FedSEA-LLaMA-Framework}
\end{figure*} 

\subsection{Federated Learning for LLMs}
Traditional Federated Learning (FL) frameworks face scalability challenges with large language models (LLMs) due to high communication and computational costs~\cite{mcmahan2017communication,stremmel2021pretraining,ji2019learning}. Recent federated LLM frameworks~\cite{fan2023fate,kuang2024federatedscope,ye2024openfedllm} address this by using instruction tuning, Adapters, and LoRA to reduce parameter updates, lowering communication overhead. However, transmission demands often still exceed system limits. To mitigate this, gradient compression~\cite{wu2024cg,shu2024ferret} and quantization~\cite{jianhao2024promoting} techniques have been applied. Structural pruning based on Fisher Information~\cite{liu2024fisher} further narrows updates to key layers, reducing model size without performance loss. Knowledge distillation~\cite{wu2024fedbiot,fan2024fedcollm,fan2025fedmkt} offers another solution, using smaller proxy models to transfer knowledge with minimal communication. Client heterogeneity remains a major challenge~\cite{zhang2024towards}, prompting methods like LoRA-based stacking~\cite{bai2024federated,wang2024flora} to align models from diverse clients and improve fairness. 

Despite advances in efficient fine-tuning and parameter reduction for federated LLMs, resource-constrained clients still face challenges with independent fine-tuning, particularly under compression overhead. Split learning~\cite{thapa2022splitfed} offers a promising alternative in such settings.

\subsection{Split Learning for LLMs}
Split learning is an emerging decentralized training approach designed for resource-constrained settings~\cite{abedi2024fedsl,rahman2020internet,matsubara2021neural}, especially mobile or GPU-limited clients. It works by dividing a neural network into sub-networks, each processed on a different device for collaborative learning. To address the training of sequential data in LLMs, FedBERT~\cite{tian2022fedbert} introduces an innovative federated learning framework designed to pre-train language models in a split architecture, addressing the challenge of limited computational capacity on client devices. Although this design enables more efficient distribution of training workloads, it comes with drawbacks, such as increased communication overhead, fixed partition points, and susceptibility to privacy threats like embedding gradient attacks.

Recently, \citet{su2024titanic} extends the scope of split learning by exploring scenarios with heterogeneous resources, particularly considering a case where the server lacks sufficient computational capacity. In their approach, the LLM is automatically divided into $k$ partitions, each assigned to one of $k$ clients, while the server merely performs aggregation. Although this strategy alleviates server-side computation, it considerably increases the overall training and inference time. Furthermore, repeated transmission of hidden states increases the risk of single points of failure and introduces potential security vulnerabilities. \citet{zheng2024safely} propose FL-GLM to place the input and output blocks locally on client devices, while the remaining primary model parameters are centralized on a server with ample computational resources. And then, they employ key encryption during client-server communication to prevent reverse engineering attacks from peer clients. But the peer-to-peer key encryption is not enough because the peer clients and servers can collaborate to infer the clients' private data.

Our work diverges from these studies by enhancing both training/inference efficiency, adaptive partition points, and data privacy for distributed LLMs through optimized algorithmic designs and communication strategies.

\section{Model}
In this section, we present a detailed description of the FedSEA-LLaMA framework, as shown in Figure~\ref{fig:FedSEA-LLaMA-Framework}, which comprises three components: privacy preservation, inference acceleration, and adaptive partitioning. Following FL-GLM, we adopt split learning to divide the LLaMA into three parts: the local client stores the first ${p}$ blocks($Blocks_A$) and the last ${q}$ blocks($Blocks_C$), while the remaining ${k}$ blocks($Blocks_B$), comprising the majority of the model parameters, are hosted on the server. 

\subsection{Privacy Preservation}
In distributed training of LLMs, the split learning paradigm partitions the entire model into multiple sub-models deployed across different physical nodes. As a result, the hidden states and gradients must be transferred across nodes via the network. Unlike traditional federated learning, which only shares gradients, this approach exposes mid-layer activations and their gradients, greatly increasing the risk of adversaries reverse-engineering user data and challenging the data protection framework of federated split LLMs.
\begin{figure}[!t]
    \centering
    \includegraphics[width=\linewidth]{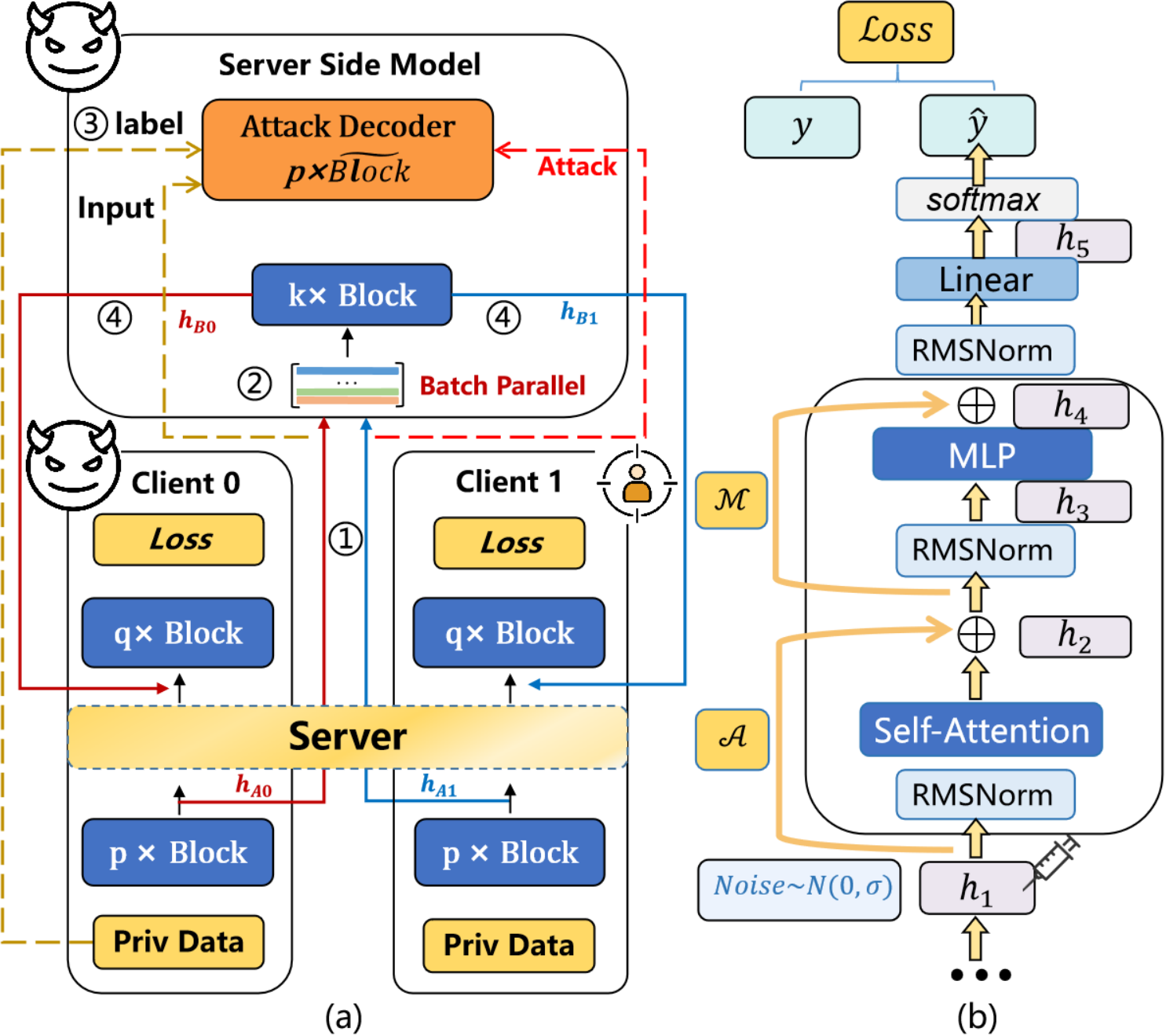}
    \caption{
    (a)Threat model attack in Multi-client Training.
    \textcircled{{\small 1}}Clients upload hidden states; 
    \textcircled{{\small 2}}Server processes via $Blocks_B$; 
    \textcircled{{\small 3}}Attack decoder learns from malicious client's data; 
    \textcircled{{\small 4}}Normal outputs are sent back to all clients.
    (b) Gaussion noise injection. Adding noise to $h_1$ provides protection for both the forward and backward passes. }
    \label{fig:Attack Process}
\end{figure}
\subsubsection{Threat Model Definition}
We design an extreme attack scenario to evaluate the robustness of FedSEA-LLaMA. Inspired by \citet{pasquini2021unleashing}, we construct a model inversion attack in a multi-client federated setting, as illustrated in Figure \ref{fig:Attack Process}a, where a malicious client colludes with an honest-but-curious server. The malicious client shares its private data in plaintext with the server, allowing the server to train an attack model locally. The goal of the attack model is to reconstruct the original input text from the hidden states transmitted by the client. During this process, the server does not tamper with the hidden states sent to other honest clients, thereby performing a stealthy inversion attack against their private data without awareness. 

Concretely, we consider an adversary that has full knowledge of the client-side model architecture and the number of layers deployed on the client, but does not have access to the actual model parameters. Thus, the honest-but-curious server locally initializes a decoder with the same architecture and depth as the client’s model and uses it as the attack model. Without loss of generality, we consider a two-client FedSEA-LLaMA setting, where the server receives hidden states from both $client_0$ and $client_1$. During each training step, the server first performs a normal forward pass through $Blocks_B$ to obtain the intermediate hidden states $h_B$. It then feeds the hidden states from the malicious $client_0$ into the attack model and uses the corresponding private data $D_{\text{priv}}$ as supervision. Afterward, $h_B$ is split and sent back to the respective clients, allowing training to proceed as usual without disrupting the federated process. 


\subsubsection{Defense with Gaussian Noise Injection}
We explore two defense strategies to protect the transmission in FedSEA-LLaMA: adding noise to gradients and adding noise to forward-pass hidden states.
The first strategy introduces random Gaussian noise into the initial gradient during backpropagation (i.e., the gradient $\nabla h_B$), while the second applies Gaussian noise to the client’s first forward-pass output (i.e., the hidden state $h_A$). Empirical results reveal that federated split LLMs are highly sensitive to gradient perturbations, so injecting noise into gradients often leads to instability in federated training, with the loss frequently diverging to NaN after several training steps. In contrast, perturbing the hidden state $h_A$ during the forward pass yields more stable training. Therefore, FedSEA-LLaMA adopts the latter to enhance robustness while preserving privacy.

Next, we derive how the noise introduced in the forward pass can also propagate to the gradients during backpropagation. Without loss of generality, we apply noise perturbation to $h_1$ in the theoretical proof, although in practice we apply noise perturbation to $h_A$. The proof will be similar in both cases. Figure \ref{fig:Attack Process}b illustrates the schematic of noise injection during the forward propagation of LLMs. Let $W_n$ denote the weight matrix of the down-projection fully-connected layer within the final MLP layer, and $W_{n+1}$ denote the weight of the output linear layer. In the noise-free scenario, the hidden states $h_1$ undergo forward propagation:
\begin{align}
    h_2 &= \mathcal{A}(h_1) +h_1, \label{Eq:h_2} \\
    h_3 &= \text{RMSNorm}(h_2), \label{Eq:h_3} \\
    h_4 &= \mathcal{M}(h_3) + h_2, \label{Eq:h_4} \\
        &=  W_n[Silu(gate_{proj}(h_3)) \cdot up_{proj}(h_3)] + h_2, \label{Eq:h_6} \\
    h_5 &=  W_{n+1} \cdot \text{RMSNorm}(h_4), \label{Eq:h_5} \\
    \hat{y} &= \text{softmax}(h_5), \\
    \mathcal{L} &= CrossEntropy(y, \hat{y}), \label{Eq:h_7}
\end{align}
where $\text{RMSNorm}(*)$ is the layernorm function, $\mathcal{A}$ is the attention layer, $\mathcal{M}$ is the MLP layer.

During backpropagation, according to \cite{clark2017computing}, the gradient of $W_n$ is:
\begin{align}
    \frac{\partial L}{\partial W_{n}} & = \frac{\partial L}{\partial h_{5}} \cdot \frac{\partial h_{5}}{\partial h_{4}} \cdot \frac{\partial h_{4}}{\partial W_{n}}, \\
    \frac{\partial L}{\partial h_{5}} &= (\hat{y} - y), \\
     \frac{\partial h_{5}}{\partial h_{4}} &= W_{n+1} \cdot J_{\text{RMSNorm}}(h_4), \\
    \frac{\partial h_{4}}{\partial W_{n}} &= [Silu(gate_{proj}(h_3)) \cdot up_{proj}(h_3)]^T. \label{Eq:h_11}
\end{align} 

According to Eq. \ref{Eq:h_2}-\ref{Eq:h_11}, the gradient of $W_n$ is closely related to the value of $h_1$. Thus, when Gaussian noise $\mathcal{N}(0,\delta)$ is injected into the hidden states $h_1$, perturbations are consequently introduced into the gradients of parameters, thereby influencing subsequent gradient computations. In other words, injecting noise into the forward propagation process can protect the hidden states and their gradients transmitted during the training of the splitting LLMs.

\subsection{Inference Acceleration}
The inference time of LLMs is crucial for the efficient and timely execution of complex tasks.
To enhance the efficiency and usability of distributed LLMs, our proposed FedSEA-LLaMA framework incorporates attention-mask compression and collaborative KV cache mechanisms to minimize redundant data transmission. 
\subsubsection{Attention-Mask Compression}
FedSEA-LLaMA transmits only the minimal information necessary for effective model training and inference. In particular, it reduces redundancy in attention-mask transmissions compared to FL-GLM. 
Since the generation of attention-mask remains consistent across transformer blocks within each training iteration and depends solely on dataset padding, transmitting full attention-mask tensors
$A_{mask}\in [0,-\infty]^{batch\_size \times seq\_len \times seq\_len}$ results in unnecessary communication overhead during training and inference. Instead, FedSEA-LLaMA transmits only essential metadata ($seq\_len$, $pad\_len$), indicating padding locations, thereby significantly decreasing bandwidth usage.

\subsubsection{Collaborative KV Cache Mechanism}
In traditional distributed inference of LLMs, each inference step involves transmitting all previously generated tokens from the client to the server. This design leads to two major inefficiencies: (1) the computational overhead increases linearly with the growing input length, since the attention mechanism must recompute key and value (KV) representations for all past tokens; and (2) the repeated transmission of the entire context between client and server introduces substantial communication latency, especially for long sequences.

To address these issues, we propose a collaborative KV cache mechanism tailored for distributed inference. Instead of recomputing and retransmitting all past tokens at each step, our approach caches the attention KV states locally on both the client and server sides. Since these KV states remain static across the autoregressive generation process, subsequent tokens can reuse the cached representations. Built on this mechanism, FedSEA-LLaMA enables highly efficient distributed inference by significantly reducing both computational and communication overhead. Experimental results show that this strategy leads to substantial improvements in inference speed in scenarios with long user queries.

\subsection{Adaptive Partition}

To flexibly accommodate specific task requirements and hardware constraints, we propose an adaptive splitting strategy that enables users to adjust the number of model input and output blocks,
thereby achieving an optimal balance between performance and efficiency. Based on this strategy, we explore the performance of downstream tasks when the client holds the first blocks (with $Blocks_A$=1,2,3) and the last blocks (with $Blocks_C$=1,2,3) of the model.
The results indicate that the depth of the model split does not significantly impact performance on downstream tasks, while affording a more flexible allocation of computational loads between the server and the clients. Furthermore, it corroborates the efficiency of our adaptive splitting strategy and delineates a refined configuration paradigm for deploying LLMs in federated learning environments.

\section{Experiments}
\label{sec:experiment}
We evaluate FedSEA-LLaMA on NLU, summarization, and conversational QA to assess its performance against centralized training while tackling latency and privacy via decentralized learning. More experiments, including parallel training and training prompts, can be found in the Appendix.

\begin{table*}[!t]
    \small
    \centering
        \begin{adjustbox}{width=\linewidth}
        \begin{tabular}{lccccccccccc}
            \toprule
             \multirow {2}*{Model}&  \textbf{ReCoRD} & \textbf{COPA} & \textbf{WSC} & \textbf{RTE} & \textbf{BoolQ} & \textbf{WiC} & \textbf{CB} & \textbf{MultiRC} & \textbf{CoQA} & \textbf{Xsum}  \\
             &F1/Acc. &Acc. &Acc. &Acc. &Acc. &Acc. & F1 &F1a/EM &F1/EM &rouge-1/rouge-2& \\  \hline
              $\text{T5}_{\text{large}}$~\cite{raffel2020exploring} & 85.7/85.0&78.0&84.6&84.8&84.3&71.6&96.4&80.9/46.6& -&40.9/17.3\\
            $\text{BART}_{\text{Large}}$~\cite{lewis2020bart} & 88.3/87.8&60.0&65.4&84.5&84.3&69.0&90.5&81.8/48.0& - & 45.1/22.3\\
            $\text{GLM}_{\text{RoBERTa}}$~\cite{glm} &  89.6/89.0&82.0&83.7&87.7&84.7&71.2&98.7&82.4/50.1& - &45.5/23.5\\
            ChatGLM-6B~\cite{zeng2022glm} &  80.2/78.7  &85.0  & 71.2 & 81.6 & 83.4 & 71.0 & 85.7 & 78.2/45.6 & - & 37.6/12.5\\
            FL-GLM~\cite{zheng2024safely}
            &79.8/78.4 & 85.0 &71.2 &80.1 &81.9 &69.6  &  85.7 &79.3/46.1 & 62.7/49.0 &37.0/11.9\\ 
            LLaMA2-7B~\cite{touvron2023LLaMA}
            &81.3/79.5 & 75.0 &68.3 &73.3 &83.8 &70.9  &  85.7 &82.3/50.1 & 88.6/80.9 &45.9/26.1\\
            \hdashline
            FedSEA-LLaMA & 81.3/79.4 &75.0 &68.4 &73.3 &82.4  &70.8 &85.7   &82.7/48.0 &88.6/80.7 &47.6/25.0 \\
            
            

            \bottomrule            
        \end{tabular}
        \end{adjustbox}   
    \caption{Performance Comparison on SupleGLUE, CoQA and Xsum datasets between FedSEA-LLaMA and baselines.}
    \label{comparison}
\end{table*}

\subsection{Experimental Settings}
In this section, we introduce some empirical settings, including datasets, evaluation metrics, baselines, and parameter settings for FedSEA-LLaMA.

\subsubsection{Dataset}
To evaluate model generalization across diverse NLP tasks, our experiments incorporate three distinct tasks:
\begin{itemize}
    \item The SuperGLUE benchmark~\cite{wang2020superglue} evaluates eight distinct NLU tasks, from common sense reasoning to fine-grained semantic interpretation. 
    
    \item The CoQA dataset~\cite{reddy2019coqa} is designed for the task of conversational question answering(QA), featuring 8k+ dialogues across seven domains. Answers are free-form text, with nearly half of the questions requiring coreference resolution and pragmatic reasoning, simulating real-world dialogue coherence.
    
    \item The XSum dataset~\cite{narayan2018don} evaluates abstractive summarization via extreme compression of BBC news articles into single-sentence summaries. It tests the model's ability to abstract and condense salient information across diverse topics.
    
\end{itemize}





\subsubsection{Metrics} 
We evaluate model performance using standard task-specific metrics. For NLU tasks such as COPA, WSC, etc, we report Accuracy, which measures the proportion of instances where the model's predicted label exactly matches the ground-truth label.

For conversational QA tasks, we use Exact Match (EM) and F1 score to evaluate model performance~\cite{yatskar2019qualitative, you2022end}. EM measures the percentage of predictions that exactly match the ground truth, accounting for punctuation and case, while F1 reflects the harmonic mean of precision and recall, rewarding partial overlaps.

For summarization tasks, we use standard ROUGE metrics~\cite{lin2004rouge,liu2021topic,fang2022spoken}, specifically ROUGE-1 and ROUGE-2, which assess unigram and bigram overlap between generated and reference summaries, capturing content coverage and fluency.

\subsubsection{Baselines} 
We evaluate FedSEA-LLaMA based on the LLaMA2-7B model~\cite{touvron2023LLaMA}, an open-source pre-trained model developed by Meta and FL-GLM. 
We additionally include several strong baselines: $\text{T5}_{\text{large}}$~\cite{raffel2020exploring}, $\text{BART}_{\text{Large}}$~\cite{lewis2020bart}, $\text{GLM}_{\text{RoBERTa}}$~\cite{glm} and $\text{ChatGLM-6B}$~\cite{glm2024chatglm}.

\subsubsection{Parameter Settings}
The server is equipped with a Gigabit Ethernet card and utilizes multiple GPUs, including two NVIDIA A6000 and two NVIDIA L40. The implementation is based on the Flower~\cite{beutel2020flower} framework, which orchestrates the transfer of intermediate data between clients and the server. Due to limited computational resources, FP32 precision and a batch size of 1 are used only for the CoQA dataset. All other experiments are conducted with FP16 precision and a batch size of 2. The random seed is consistently set to 42. Fine-tuning is performed using distributed LoRA training, where $Blocks_A$, $Blocks_B$, and $Blocks_C$ each maintain their own LoRA adapters. For all LoRA fine-tuning experiments in this paper, the target modules are set to q\_proj, v\_proj, k\_proj, and o\_proj.

\subsection{Experimental Results}
\subsubsection{FedSEA-LLaMA Performance}
\begin{figure}[!t]
\centering
\includegraphics[width=0.8\linewidth]{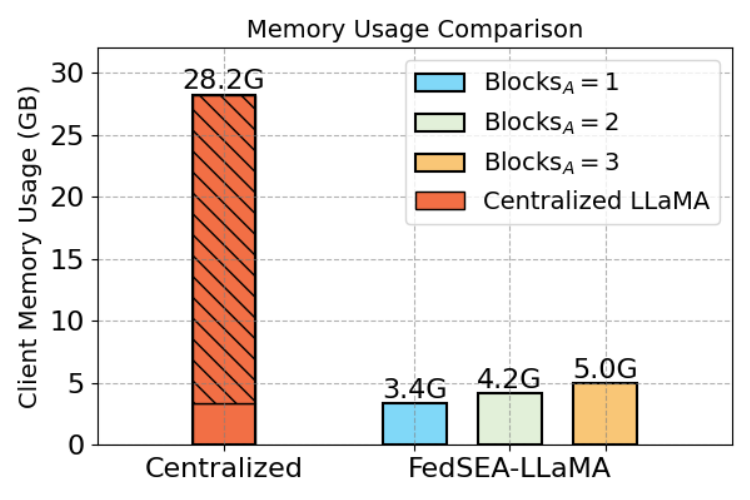}
\caption{Computation load on client side of centralized LLaMA2-7B and FedSEA-LLaMA on ReCoRD dataset.}
\label{fig:gpu_comparison}
\end{figure}

We compared the performance of FedSEA-LLaMA with centralized LLaMA2-7B on various tasks, including NLU, summarization generation, and conversational QA, as shown in Table~\ref{comparison}. 
Experimental results show that FedSEA-LLaMA achieves comparable performance to LLaMA2-7B, with metrics approaching or matching across multiple tasks, demonstrating its effectiveness. Meanwhile, compared to other baselines, FedSEA-LLaMA achieves better performance on datasets such as Record, MultiRC, and Xsum, but performs poorly on the RTE dataset. This is speculated to be related to the distribution difference between RTE and the data used for LLaMA2 pre-training, as well as suboptimal hyperparameters.
The comparison of their GPU memory usage is shown in Figure~\ref{fig:gpu_comparison}. The result shows that FedSEA-LLaMA significantly reduces GPU memory consumption on the client side. Both models were fine-tuned on the Record dataset with a batch size of 2, using float16 precision and a LoRA rank of 8. While centralized LLaMA2-7B occupies 28.2G of GPU memory, FedSEA-LLaMA client requires only 3.4G, representing a reduction of up to 87.9\% in memory requirements.

\begin{figure*}[!t] 
\centering
\captionsetup[subfigure]{justification=centering}
\begin{subfigure}[b]{0.32\linewidth}
    \includegraphics[width=\linewidth]{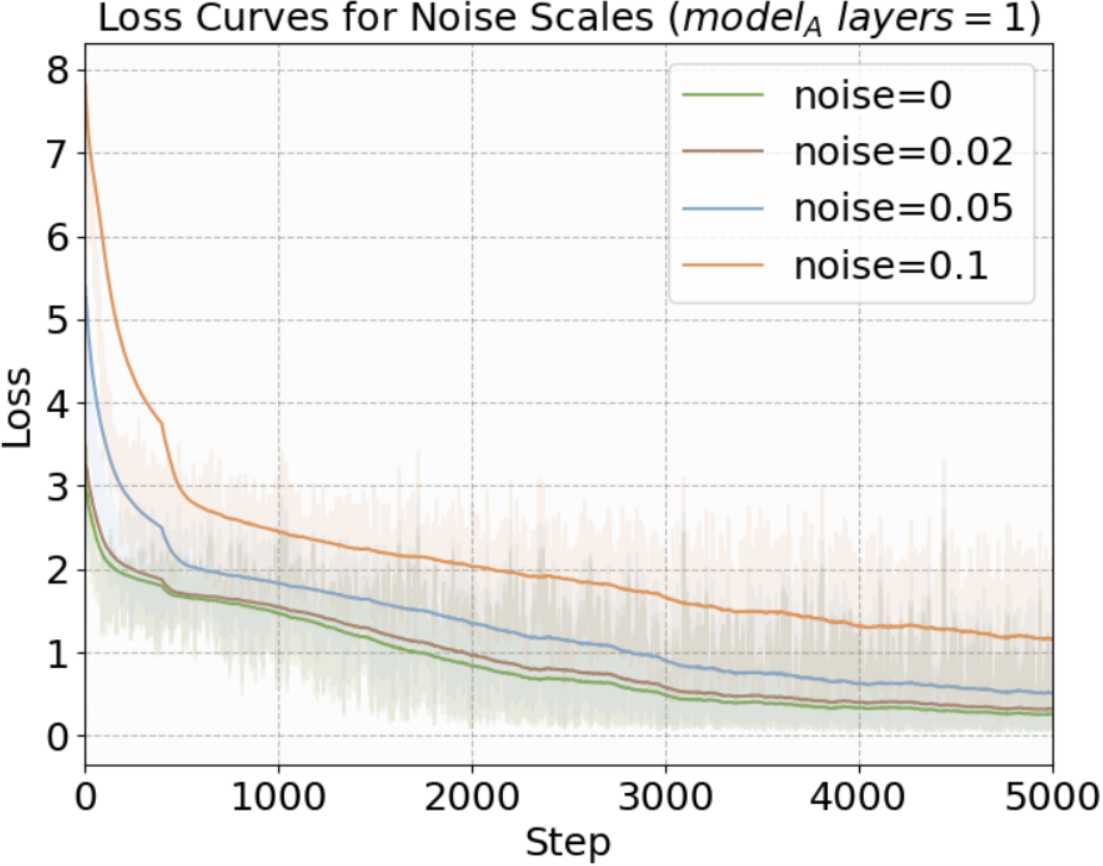}
    \caption{Loss Curves for Noise Scales \newline($Blocks_A$=1, $Blocks_C$=1)}
    \label{fig:Loss_for_Blocks_A_1}
\end{subfigure}
\hfill 
\begin{subfigure}[b]{0.33\linewidth}
    \includegraphics[width=\linewidth]{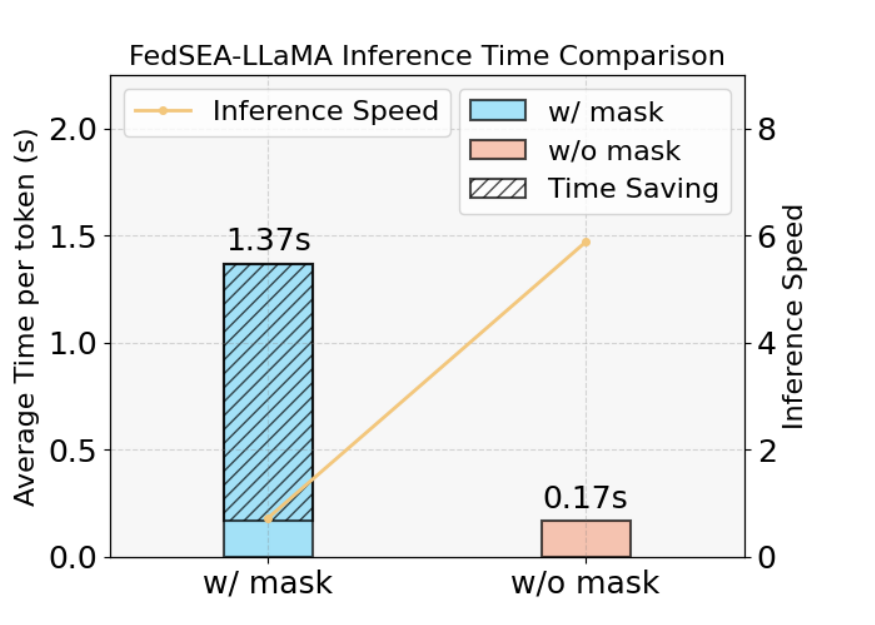}
    \caption{Inference Speed Comparison \\(w/ vs. w/o Attention-Mask).}
    \label{fig:w_wo_mask}
\end{subfigure}
\hfill 
\begin{subfigure}[b]{0.32\linewidth}
    \includegraphics[width=\linewidth]{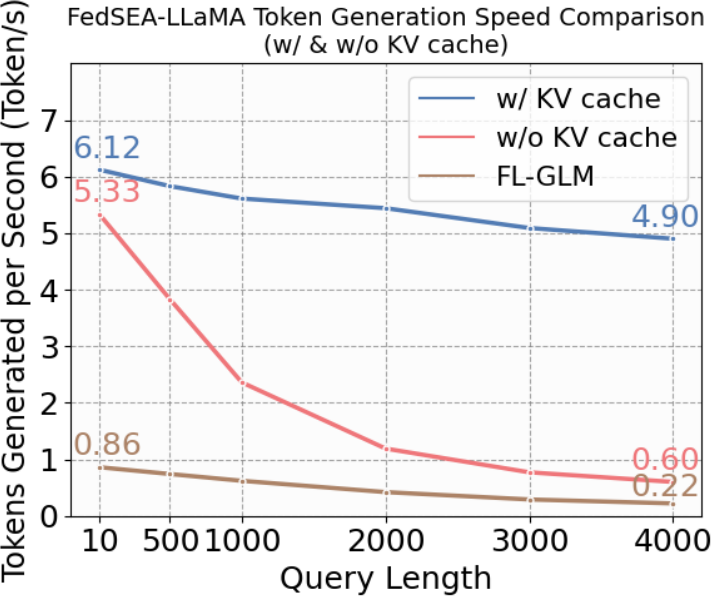}
    \caption{Inference Speed Comparison \\(w/ vs. w/o KV Cache).}
    \label{fig:kvcache}
\end{subfigure}

\caption{(a)Different noise injection on MultiRC dataset. (b)Ablation study on attention-mask. (c)Ablation study on KV cache.}

\end{figure*}

\subsubsection{Effects of Noise Scales} \label{sec:Noise injection}
                                
Since noise injection into the hidden states during forward propagation inevitably impacts the performance of distributed LLMs, we conducted extensive experiments to investigate how noise levels affect model performance. The corresponding experimental results are presented in Figure \ref{fig:Loss_for_Blocks_A_1} and Table~\ref{tab:performance_modelA_noise}. 
Figure \ref{fig:Loss_for_Blocks_A_1} demonstrates that when the numbers of blocks in $Blocks_A$ and $Blocks_C$ are consistent, the convergence speed of FedSEA-LLaMA decreases as the magnitude of injected noise increases. When the average absolute value of the noise is small (e.g., below 0.02), the loss curve remains close to that of the noise-free baseline.
Table~\ref{tab:performance_modelA_noise} indicates that under fixed noise levels, increasing the number of blocks in $Blocks_A$ consistently improves downstream performance in terms of both F1 and EM scores. Notably, when the noise amplitude reaches 0.1, only the single-block configuration of $Blocks_A$ shows a significant drop in downstream performance, whereas the three-block version of $Blocks_A$ demonstrates greater robustness.

\begin{table}[!t]
\centering
\begin{adjustbox}{width=\linewidth}
\begin{tabular}{cccc}
\toprule
\textbf{Noise Scale} & \textbf{$Blocks_A$ = 1} & \textbf{$Blocks_A$ = 2} & \textbf{$Blocks_A$ = 3} \\
\hline
0.0   & 77.48/40.82 & 77.82/39.24 & 77.89/41.34 \\
0.02  & 77.89/38.09 & 77.87/39.66 & 78.61/42.39 \\
0.05  & 74.57/29.17 & 77.07/38.41 & 78.95/41.03\\
0.1   & 59.11/8.18 & 72.45/23.71 & 75.68/34.10 \\
\bottomrule

\end{tabular}
\end{adjustbox} 
\caption{Different noise scales and blocks number of $Blocks_A$ analysis(F1a / EM) on MultiRC dataset.}
\label{tab:performance_modelA_noise}
\end{table}

\subsubsection{Model Reverse Attack}\label{sec:Model Reverse Attack}
To assess how split points and noise scale affect the security of FedSEA-LLaMA, 
we evaluate the attack performance of the adversarial model under varying configurations of $Blocks_A$ depth ($Blocks_A$ = 1, 2, 3) and noise (scale = 0, 0.02, 0.5). 
Attacks use full-parameter fine-tuning (lr=2e-5), with reconstruction quality measured by BLEU-4 and ROUGE-2. 
FedBert reports a Rouge-2 of 7.053 and a Bleu-4 of 28.57. Experiment results of FedSEA-LLaMA are shown in Table~ \ref{tab:Attack_performance_modelA_noise}. From the results, we can see that when the client holds only the embedding layer (similar to FedBERT), the attack model achieves relatively high BLEU and ROUGE scores. However, under the FedSEA-LLaMA framework, where the client also retains a portion of the LLM blocks in addition to the embedding layer, all attack metrics approach zero. 
These findings demonstrate that FedSEA-LLaMA effectively mitigates model inversion attacks, thereby preserving the privacy of client-side data.


\begin{table}[!t]
\centering
\begin{adjustbox}{width=\linewidth}
\begin{tabular}{cccc}
\toprule
\textbf{Noise scale} & \textbf{$Blocks_A$=1} & \textbf{$Blocks_A$=2} & \textbf{$Blocks_A$=3}\\
\hline
0.0   & 0.051/0.003 & 0.056/0.003 & 0.069/0.016 \\
0.02  & 0.050/0.002 & 0.052/0.003 & 0.084/0.044 \\
0.05  & 0.050/0.003 & 0.051/0.002 & 0.067/0.004 \\
\bottomrule
\end{tabular}
\end{adjustbox}
\caption{Security analysis (Rouge-2 F1 / BlEU-4) of FedSEA-LLaMA on Record dataset.}
\label{tab:Attack_performance_modelA_noise}
\end{table}



\subsubsection{Efficiency Optimization} 
Figure~\ref{fig:w_wo_mask} shows a comparison of the time consumption between transferring the attention-mask and transferring only the actual input length during distributed training. The bar chart represents the average time spent per token in FedSEA-LLaMA's forward propagation, while the line chart represents the number of tokens generated per second(context length=500). We can see that transferring the compressed information of the attention-mask plays a significant role in reducing the latency of distributed LLMs, with an average time reduction of 87.6\%.


As shown in Figure~\ref{fig:kvcache}, FedSEA-LLaMA demonstrates superior performance compared to FL-GLM in distributed inference, achieving over 7× speedup. Without the KV cache, the inference speed decreases significantly with longer queries, due to the increasing number of K and V vectors. In contrast, with the KV cache technique, the client only needs to send the intermediate vector of the current single token to the server, while reusing previously generated K and V vectors. This effectively avoids redundant computations. When the query length reaches 4000, the inference speed of FedSEA-LLaMA with KV cache is 8.2 times that without it. As query length increases further, this advantage continues to expand, greatly alleviating the computational burden and operational latency of distributed LLMs.

\subsubsection{Adaptive Splitting} \label{sec:adaptive}
The choice of model partitioning has a significant impact on the computational burden at the client side. To evaluate whether FedSEA-LLaMA maintains robust performance under different partitioning strategies, we investigate its downstream task performance when clients hold different numbers of blocks in $Blocks_A$ and $Blocks_C$ ($Blocks_A$ = 1, 2, 3 and $Blocks_C$ = 1, 2, 3). This setting reflects realistic constraints where clients possess limited computational resources. 
Table~\ref{tab:em_partition_noise_free} shows F1a/EM scores across different partitioning schemes.
Performance variations across partitions are minimal, indicating that FedSEA-LLaMA is robust to different split configurations. This flexibility allows resource-constrained clients to adjust block allocation based on their downstream task requirements and resource availability, without significant accuracy loss.

\begin{table}[!t]
\centering
\begin{adjustbox}{width=\linewidth}
\begin{tabular}{cccc}
\toprule
\textbf{$Blocks_A$ } & \textbf{$Blocks_C$ = 1} & \textbf{$Blocks_C$ = 2} & \textbf{$Blocks_C$ = 3} \\
\hline
1 & 80.34/44.60 & 80.63/46.59 & 80.90/47.11 \\
2 & 80.41/45.86 & 80.62/45.96 & 79.94/45.02 \\
3 & 81.17/47.22 & 80.74/46.59 & 80.20/45.33 \\
\toprule
\end{tabular}
\end{adjustbox}
\caption{ Different model partition analysis(F1a/EM) on MultiRC dataset. No Noise is added to the hidden states.}
\label{tab:em_partition_noise_free}
\end{table}


\section{Conclusion} \label{sec:conclusion}
In this work, we propose FedSEA-LLaMA, a novel federated split learning framework tailored for LLMs. By securely offloading the majority of model computation to a central server while maintaining privacy-sensitive components on local clients, FedSEA-LLaMA addresses the core limitations of traditional federated split learning approaches. Through secure end-to-end propagation with noise injection, inference acceleration, and adaptive partitioning tailored to downstream tasks and hardware, FedSEA-LLaMA demonstrates significant advances in security, efficiency, and adaptability. 
This federated split LLMs framework not only keeps users' private data confined to their local devices but also significantly shortens training and inference time, making it better suited for handling the scale and complexity of LLMs. In future work, we aim to support more base models and deploy the FedSEA-LLaMA framework in scenarios where data privacy is a critical concern.

\section*{Acknowledgements}
This work was funded by the National Natural Science Foundation of China (NSFC) under Grants No. 62406013,No.62388101, the Beijing Advanced Innovation
Center Funds for Future Blockchain and Privacy Computing(GJJ-24-034), and the Fundamental Research Funds for the Central Universities.

\bibliography{aaai2026}

@Misc{		  wang2020superglue,
  title		= {SuperGLUE: A Stickier Benchmark for General-Purpose
		  Language Understanding Systems},
  author	= {Alex Wang and Yada Pruksachatkun and Nikita Nangia and
		  Amanpreet Singh and Julian Michael and Felix Hill and Omer
		  Levy and Samuel R. Bowman},
  year		= {2020},
  eprint	= {1905.00537},
  archiveprefix	= {arXiv},
  primaryclass	= {cs.CL},
  url		= {https://arxiv.org/abs/1905.00537}
}

@InProceedings{	  yatskar2019qualitative,
  title		= {A Qualitative Comparison of CoQA, SQuAD 2.0 and QuAC},
  author	= {Yatskar, Mark},
  booktitle	= {Proceedings of the 2019 Conference of the North American
		  Chapter of the Association for Computational Linguistics:
		  Human Language Technologies, Volume 1 (Long and Short
		  Papers)},
  pages		= {2318--2323},
  year		= {2019}
}

@InProceedings{	  you2022end,
  title		= {End-to-end Spoken Conversational Question Answering: Task,
		  Dataset and Model},
  author	= {You, Chenyu and Chen, Nuo and Liu, Fenglin and Ge, Shen
		  and Wu, Xian and Zou, Yuexian},
  booktitle	= {Findings of the Association for Computational Linguistics:
		  NAACL 2022},
  pages		= {1219--1232},
  year		= {2022}
}

@Article{	  reddy2019coqa,
  title		= {Coqa: A conversational question answering challenge},
  author	= {Reddy, Siva and Chen, Danqi and Manning, Christopher D},
  journal	= {Transactions of the Association for Computational
		  Linguistics},
  volume	= {7},
  pages		= {249--266},
  year		= {2019},
  publisher	= {MIT Press One Rogers Street, Cambridge, MA 02142-1209, USA
		  journals-info~…}
}

@article{glm2024chatglm,
  title={Chatglm: A family of large language models from glm-130b to glm-4 all tools},
  author={GLM, Team and Zeng, Aohan and Xu, Bin and Wang, Bowen and Zhang, Chenhui and Yin, Da and Zhang, Dan and Rojas, Diego and Feng, Guanyu and Zhao, Hanlin and others},
  journal={arXiv preprint arXiv:2406.12793},
  year={2024}
}

@Article{	  touvron2023llama,
  title		= {Llama 2: Open foundation and fine-tuned chat models},
  author	= {Touvron, Hugo and Martin, Louis and Stone, Kevin and
		  Albert, Peter and Almahairi, Amjad and Babaei, Yasmine and
		  Bashlykov, Nikolay and Batra, Soumya and Bhargava, Prajjwal
		  and Bhosale, Shruti and others},
  journal	= {arXiv preprint arXiv:2307.09288},
  year		= {2023}
}

@Article{	  raffel2020exploring,
  title		= {Exploring the limits of transfer learning with a unified
		  text-to-text transformer},
  author	= {Raffel, Colin and Shazeer, Noam and Roberts, Adam and Lee,
		  Katherine and Narang, Sharan and Matena, Michael and Zhou,
		  Yanqi and Li, Wei and Liu, Peter J},
  journal	= {Journal of machine learning research},
  volume	= {21},
  number	= {140},
  pages		= {1--67},
  year		= {2020}
}

@Article{	  beutel2020flower,
  title		= {Flower: A Friendly Federated Learning Research Framework},
  author	= {Beutel, Daniel J and Topal, Taner and Mathur, Akhil and
		  Qiu, Xinchi and Fernandez-Marques, Javier and Gao, Yan and
		  Sani, Lorenzo and Kwing, Hei Li and Parcollet, Titouan and
		  Gusmão, Pedro PB de and Lane, Nicholas D},
  journal	= {arXiv preprint arXiv:2007.14390},
  year		= {2020}
}

@InProceedings{	  pasquini2021unleashing,
  title		= {Unleashing the tiger: Inference attacks on split
		  learning},
  author	= {Pasquini, Dario and Ateniese, Giuseppe and Bernaschi,
		  Massimo},
  booktitle	= {Proceedings of the 2021 ACM SIGSAC Conference on Computer
		  and Communications Security},
  pages		= {2113--2129},
  year		= {2021}
}

@inproceedings{lin2004rouge,
  title={Rouge: A package for automatic evaluation of summaries},
  author={Lin, Chin-Yew},
  booktitle={Text summarization branches out},
  pages={74--81},
  year={2004}
}

@InProceedings{	  zheng2024safely,
  title		= {Safely Learning with Private Data: A Federated Learning
		  Framework for Large Language Model},
  author	= {Zheng, Jia-Ying and Zhang, Hainan and Wang, Lingxiang and
		  Qiu, Wangjie and Zheng, Hong-Wei and Zheng, Zhi-Ming},
  booktitle	= {Proceedings of the 2024 Conference on Empirical Methods in
		  Natural Language Processing},
  pages		= {5293--5306},
  year		= {2024}
}

@InProceedings{	  su2024titanic,
  title		= {TITANIC: Towards production federated learning with large
		  language models},
  author	= {Su, Ningxin and Hu, Chenghao and Li, Baochun and Li, Bo},
  booktitle	= {IEEE INFOCOM 2024-IEEE Conference on Computer
		  Communications},
  pages		= {611--620},
  year		= {2024},
  organization	= {IEEE}
}

@Article{	  fan2023fate,
  title		= {Fate-llm: A industrial grade federated learning framework
		  for large language models},
  author	= {Fan, Tao and Kang, Yan and Ma, Guoqiang and Chen, Weijing
		  and Wei, Wenbin and Fan, Lixin and Yang, Qiang},
  journal	= {arXiv preprint arXiv:2310.10049},
  year		= {2023}
}

@InProceedings{	  kuang2024federatedscope,
  title		= {Federatedscope-llm: A comprehensive package for
		  fine-tuning large language models in federated learning},
  author	= {Kuang, Weirui and Qian, Bingchen and Li, Zitao and Chen,
		  Daoyuan and Gao, Dawei and Pan, Xuchen and Xie, Yuexiang
		  and Li, Yaliang and Ding, Bolin and Zhou, Jingren},
  booktitle	= {Proceedings of the 30th ACM SIGKDD Conference on Knowledge
		  Discovery and Data Mining},
  pages		= {5260--5271},
  year		= {2024}
}

@InProceedings{	  ye2024openfedllm,
  title		= {Openfedllm: Training large language models on
		  decentralized private data via federated learning},
  author	= {Ye, Rui and Wang, Wenhao and Chai, Jingyi and Li, Dihan
		  and Li, Zexi and Xu, Yinda and Du, Yaxin and Wang, Yanfeng
		  and Chen, Siheng},
  booktitle	= {Proceedings of the 30th ACM SIGKDD conference on knowledge
		  discovery and data mining},
  pages		= {6137--6147},
  year		= {2024}
}

@Article{	  wu2024cg,
  title		= {CG-FedLLM: How to Compress Gradients in Federated
		  Fune-tuning for Large Language Models},
  author	= {Wu, Huiwen and Li, Xiaohan and Zhang, Deyi and Xu,
		  Xiaogang and Wu, Jiafei and Zhao, Puning and Liu, Zhe},
  journal	= {CoRR},
  year		= {2024}
}

@InProceedings{	  shu2024ferret,
  title		= {Ferret: Federated Full-Parameter Tuning at Scale for Large
		  Language Models},
  author	= {Shu, Yao and Hu, Wenyang and Ng, See-Kiong and Low, Bryan
		  Kian Hsiang and Yu, Fei Richard},
  booktitle	= {International Workshop on Federated Foundation Models in
		  Conjunction with NeurIPS 2024},
  year		= {2024}
}

@InProceedings{	  jianhao2024promoting,
  title		= {Promoting Data and Model Privacy in Federated Learning
		  through Quantized LoRA},
  author	= {JianHao, Zhu and Lv, Changze and Wang, Xiaohua and Wu,
		  Muling and Liu, Wenhao and Li, Tianlong and Ling, Zixuan
		  and Zhang, Cenyuan and Zheng, Xiaoqing and Huang,
		  Xuan-Jing},
  booktitle	= {Findings of the Association for Computational Linguistics:
		  EMNLP 2024},
  pages		= {10501--10512},
  year		= {2024}
}

@InProceedings{	  wu2024fedbiot,
  title		= {Fedbiot: Llm local fine-tuning in federated learning
		  without full model},
  author	= {Wu, Feijie and Li, Zitao and Li, Yaliang and Ding, Bolin
		  and Gao, Jing},
  booktitle	= {Proceedings of the 30th ACM SIGKDD Conference on Knowledge
		  Discovery and Data Mining},
  pages		= {3345--3355},
  year		= {2024}
}

@Article{	  fan2024fedcollm,
  title		= {FedCoLLM: A Parameter-Efficient Federated Co-tuning
		  Framework for Large and Small Language Models},
  author	= {Fan, Tao and Kang, Yan and Ma, Guoqiang and Fan, Lixin and
		  Chen, Kai and Yang, Qiang},
  journal	= {arXiv preprint arXiv:2411.11707},
  year		= {2024}
}

@InProceedings{	  fan2025fedmkt,
  title		= {FedMKT: Federated Mutual Knowledge Transfer for Large and
		  Small Language Models},
  author	= {Fan, Tao and Ma, Guoqiang and Kang, Yan and Gu, Hanlin and
		  Song, Yuanfeng and Fan, Lixin and Chen, Kai and Yang,
		  Qiang},
  booktitle	= {Proceedings of the 31st International Conference on
		  Computational Linguistics},
  pages		= {243--255},
  year		= {2025}
}

@InProceedings{	  liu2024fisher,
  title		= {Fisher Information-based Efficient Curriculum Federated
		  Learning with Large Language Models},
  author	= {Liu, Ji and Ren, Jiaxiang and Jin, Ruoming and Zhang,
		  Zijie and Zhou, Yang and Valduriez, Patrick and Dou,
		  Dejing},
  booktitle	= {Proceedings of the 2024 Conference on Empirical Methods in
		  Natural Language Processing},
  pages		= {10497--10523},
  year		= {2024}
}

@Article{	  bai2024federated,
  title		= {Federated fine-tuning of large language models under
		  heterogeneous language tasks and client resources},
  author	= {Bai, Jiamu and Chen, Daoyuan and Qian, Bingchen and Yao,
		  Liuyi and Li, Yaliang},
  journal	= {arXiv e-prints},
  pages		= {arXiv--2402},
  year		= {2024}
}

@Article{	  wang2024flora,
  title		= {FLoRA: Federated Fine-Tuning Large Language Models with
		  Heterogeneous Low-Rank Adaptations},
  author	= {Wang, Ziyao and Shen, Zheyu and He, Yexiao and Sun,
		  Guoheng and Wang, Hongyi and Lyu, Lingjuan and Li, Ang},
  journal	= {CoRR},
  year		= {2024}
}

@InProceedings{	  zhang2024towards,
  title		= {Towards building the federatedgpt: Federated instruction
		  tuning},
  author	= {Zhang, Jianyi and Vahidian, Saeed and Kuo, Martin and Li,
		  Chunyuan and Zhang, Ruiyi and Yu, Tong and Wang, Guoyin and
		  Chen, Yiran},
  booktitle	= {ICASSP 2024-2024 IEEE International Conference on
		  Acoustics, Speech and Signal Processing (ICASSP)},
  pages		= {6915--6919},
  year		= {2024},
  organization	= {IEEE}
}

@inproceedings{mcmahan2017communication,
  title={Communication-efficient learning of deep networks from decentralized data},
  author={McMahan, Brendan and Moore, Eider and Ramage, Daniel and Hampson, Seth and y Arcas, Blaise Aguera},
  booktitle={Artificial intelligence and statistics},
  pages={1273--1282},
  year={2017},
  organization={PMLR}
}

@InProceedings{	  stremmel2021pretraining,
  title		= {Pretraining federated text models for next word
		  prediction},
  author	= {Stremmel, Joel and Singh, Arjun},
  booktitle	= {Advances in Information and Communication: Proceedings of
		  the 2021 Future of Information and Communication Conference
		  (FICC), Volume 2},
  pages		= {477--488},
  year		= {2021},
  organization	= {Springer}
}

@InProceedings{	  ji2019learning,
  title		= {Learning private neural language modeling with attentive
		  aggregation},
  author	= {Ji, Shaoxiong and Pan, Shirui and Long, Guodong and Li,
		  Xue and Jiang, Jing and Huang, Zi},
  booktitle	= {2019 International joint conference on neural networks
		  (IJCNN)},
  pages		= {1--8},
  year		= {2019},
  organization	= {IEEE}
}

@InProceedings{	  thapa2022splitfed,
  title		= {Splitfed: When federated learning meets split learning},
  author	= {Thapa, Chandra and Arachchige, Pathum Chamikara Mahawaga
		  and Camtepe, Seyit and Sun, Lichao},
  booktitle	= {Proceedings of the AAAI Conference on Artificial
		  Intelligence},
  volume	= {36},
  pages		= {8485--8493},
  year		= {2022}
}

@Article{	  tian2022fedbert,
  title		= {FedBERT: When federated learning meets pre-training},
  author	= {Tian, Yuanyishu and Wan, Yao and Lyu, Lingjuan and Yao,
		  Dezhong and Jin, Hai and Sun, Lichao},
  journal	= {ACM Transactions on Intelligent Systems and Technology
		  (TIST)},
  volume	= {13},
  number	= {4},
  pages		= {1--26},
  year		= {2022},
  publisher	= {ACM New York, NY}
}

@Article{	  abedi2024fedsl,
  title		= {FedSL: Federated split learning on distributed sequential
		  data in recurrent neural networks},
  author	= {Abedi, Ali and Khan, Shehroz S},
  journal	= {Multimedia Tools and Applications},
  volume	= {83},
  number	= {10},
  pages		= {28891--28911},
  year		= {2024},
  publisher	= {Springer}
}

@Article{	  rahman2020internet,
  title		= {Internet of things intrusion detection: Centralized,
		  on-device, or federated learning?},
  author	= {Rahman, Sawsan Abdul and Tout, Hanine and Talhi,
		  Chamseddine and Mourad, Azzam},
  journal	= {IEEE Network},
  volume	= {34},
  number	= {6},
  pages		= {310--317},
  year		= {2020},
  publisher	= {IEEE}
}

@InProceedings{	  matsubara2021neural,
  title		= {Neural compression and filtering for edge-assisted
		  real-time object detection in challenged networks},
  author	= {Matsubara, Yoshitomo and Levorato, Marco},
  booktitle	= {2020 25th International Conference on Pattern Recognition
		  (ICPR)},
  pages		= {2272--2279},
  year		= {2021},
  organization	= {IEEE}
}

@InProceedings{	  lewis2020bart,
  title		= {BART: Denoising Sequence-to-Sequence Pre-training for
		  Natural Language Generation, Translation, and
		  Comprehension},
  author	= {Lewis, Mike and Liu, Yinhan and Goyal, Naman and
		  Ghazvininejad, Marjan and Mohamed, Abdelrahman and Levy,
		  Omer and Stoyanov, Veselin and Zettlemoyer, Luke},
  booktitle	= {Proceedings of the 58th Annual Meeting of the Association
		  for Computational Linguistics},
  pages		= {7871--7880},
  year		= {2020}
}

@Article{	  zeng2022glm,
  title		= {Glm-130b: An open bilingual pre-trained model},
  author	= {Zeng, Aohan and Liu, Xiao and Du, Zhengxiao and Wang,
		  Zihan and Lai, Hanyu and Ding, Ming and Yang, Zhuoyi and
		  Xu, Yifan and Zheng, Wendi and Xia, Xiao and others},
  journal	= {arXiv preprint arXiv:2210.02414},
  year		= {2022}
}

@InProceedings{	  glm,
  title		= {GLM: General Language Model Pretraining with
		  Autoregressive Blank Infilling},
  author	= {Du, Zhengxiao and Qian, Yujie and Liu, Xiao and Ding, Ming
		  and Qiu, Jiezhong and Yang, Zhilin and Tang, Jie},
  booktitle	= {Proceedings of the 60th Annual Meeting of the Association
		  for Computational Linguistics (Volume 1: Long Papers)},
  pages		= {320--335},
  year		= {2022}
}

@InProceedings{	  narayan2018don,
  title		= {Don’t Give Me the Details, Just the Summary! Topic-Aware
		  Convolutional Neural Networks for Extreme Summarization},
  author	= {Narayan, Shashi and Cohen, Shay B and Lapata, Mirella},
  booktitle	= {Proceedings of the 2018 Conference on Empirical Methods in
		  Natural Language Processing},
  pages		= {1797--1807},
  year		= {2018}
}

@InProceedings{	  liu2021topic,
  title		= {Topic-Aware Contrastive Learning for Abstractive Dialogue
		  Summarization},
  author	= {Liu, Junpeng and Zou, Yanyan and Zhang, Hainan and Chen,
		  Hongshen and Ding, Zhuoye and Yuan, Caixia and Wang,
		  Xiaojie},
  booktitle	= {Findings of the Association for Computational Linguistics:
		  EMNLP 2021},
  pages		= {1229--1243},
  year		= {2021}
}

@InProceedings{	  fang2022spoken,
  title		= {From spoken dialogue to formal summary: An utterance
		  rewriting for dialogue summarization},
  author	= {Fang, Yue and Zhang, Hainan and Chen, Hongshen and Ding,
		  Zhuoye and Long, Bo and Lan, Yanyan and Zhou, Yanquan},
  booktitle	= {Proceedings of the 2022 Conference of the North American
		  Chapter of the Association for Computational Linguistics:
		  Human Language Technologies},
  pages		= {3859--3869},
  year		= {2022}
}

@Article{	  asnani2023reverse,
  title		= {Reverse engineering of generative models: Inferring model
		  hyperparameters from generated images},
  author	= {Asnani, Vishal and Yin, Xi and Hassner, Tal and Liu,
		  Xiaoming},
  journal	= {IEEE Transactions on Pattern Analysis and Machine
		  Intelligence},
  year		= {2023},
  publisher	= {IEEE}
}

@Article{	  kaplan2020scaling,
  title		= {Scaling laws for neural language models},
  author	= {Kaplan, Jared and McCandlish, Sam and Henighan, Tom and
		  Brown, Tom B and Chess, Benjamin and Child, Rewon and Gray,
		  Scott and Radford, Alec and Wu, Jeffrey and Amodei, Dario},
  journal	= {arXiv preprint arXiv:2001.08361},
  year		= {2020}
}

@Article{	  yang2024federated,
  title		= {Federated continual learning via knowledge fusion: A
		  survey},
  author	= {Yang, Xin and Yu, Hao and Gao, Xin and Wang, Hao and
		  Zhang, Junbo and Li, Tianrui},
  journal	= {IEEE Transactions on Knowledge and Data Engineering},
  volume	= {36},
  number	= {8},
  pages		= {3832--3850},
  year		= {2024},
  publisher	= {IEEE}
}

@Article{	  huang2024federated,
  title		= {Federated learning for generalization, robustness,
		  fairness: A survey and benchmark},
  author	= {Huang, Wenke and Ye, Mang and Shi, Zekun and Wan,
		  Guancheng and Li, He and Du, Bo and Yang, Qiang},
  journal	= {IEEE Transactions on Pattern Analysis and Machine
		  Intelligence},
  year		= {2024},
  publisher	= {IEEE}
}

@Article{	  yuan2024decentralized,
  title		= {Decentralized federated learning: A survey and
		  perspective},
  author	= {Yuan, Liangqi and Wang, Ziran and Sun, Lichao and Yu,
		  Philip S and Brinton, Christopher G},
  journal	= {IEEE Internet of Things Journal},
  year		= {2024},
  publisher	= {IEEE}
}

@Article{	  chen2024federated,
  title		= {When federated learning meets privacy-preserving
		  computation},
  author	= {Chen, Jingxue and Yan, Hang and Liu, Zhiyuan and Zhang,
		  Min and Xiong, Hu and Yu, Shui},
  journal	= {ACM Computing Surveys},
  volume	= {56},
  number	= {12},
  pages		= {1--36},
  year		= {2024},
  publisher	= {ACM New York, NY}
}

@Article{	  ghimire2022recent,
  title		= {Recent advances on federated learning for cybersecurity
		  and cybersecurity for federated learning for internet of
		  things},
  author	= {Ghimire, Bimal and Rawat, Danda B},
  journal	= {IEEE Internet of Things Journal},
  volume	= {9},
  number	= {11},
  pages		= {8229--8249},
  year		= {2022},
  publisher	= {IEEE}
}

@Article{	  clark2017computing,
  title		= {Computing neural network gradients},
  author	= {Clark, Kevin},
  year		= {2017},
  publisher	= {unpublished},
  journal	= {unpublished Journal}
}

@Article{	  hu2022lora,
  title		= {Lora: Low-rank adaptation of large language models.},
  author	= {Hu, Edward J and Shen, Yelong and Wallis, Phillip and
		  Allen-Zhu, Zeyuan and Li, Yuanzhi and Wang, Shean and Wang,
		  Lu and Chen, Weizhu and others},
  journal	= {ICLR},
  volume	= {1},
  number	= {2},
  pages		= {3},
  year		= {2022}
}

@InProceedings{	  houlsby2019parameter,
  title		= {Parameter-efficient transfer learning for NLP},
  author	= {Houlsby, Neil and Giurgiu, Andrei and Jastrzebski,
		  Stanislaw and Morrone, Bruna and De Laroussilhe, Quentin
		  and Gesmundo, Andrea and Attariyan, Mona and Gelly,
		  Sylvain},
  booktitle	= {International conference on machine learning},
  pages		= {2790--2799},
  year		= {2019},
  organization	= {PMLR}
}

@Article{	  hinton2015distilling,
  title		= {Distilling the knowledge in a neural network},
  author	= {Hinton, Geoffrey and Vinyals, Oriol and Dean, Jeff},
  journal	= {arXiv preprint arXiv:1503.02531},
  year		= {2015}
}

\end{document}